\documentclass{article}
\usepackage[margin=3cm]{geometry}

\usepackage[hyphens]{url}  
\usepackage{graphicx} 

\usepackage{caption} 

\usepackage{algorithm}
\usepackage{algpseudocode}

\usepackage{amsmath}
\usepackage{amssymb}
\usepackage{amsthm}
\usepackage{multirow}
\usepackage{wrapfig}
\usepackage{graphicx}
\usepackage{booktabs}
\usepackage[dvipsnames]{xcolor}

\title{Cyclical Pruning for Sparse Neural Networks}
\author{
    Suraj Srinivas$^1$\footnote{Work done while on internship at Qualcomm. Email: suuraj.srinivas@gmail.com}
    \and Andrey Kuzmin$^2$ 
    \and Markus Nagel$^2$
    \and Mart van Baalen$^2$
    \and Andrii Skliar$^2$
    \and Tijmen Blankevoort$^2$
}
\date{$^1$ Idiap Research Institute \& EPFL, Switzerland \\
      $^2$ Qualcomm AI Research, Netherlands}

\newtheorem{obs}{Observation}
\newtheorem*{defn}{Definition}

\begin{document}

\maketitle

\begin{abstract}
Current methods for pruning neural network weights iteratively apply magnitude-based pruning on
the model weights and re-train the resulting model to recover lost accuracy. In this work, we show that such strategies do not allow for the recovery of erroneously pruned weights. To enable weight recovery, we propose a simple strategy called \textit{cyclical pruning} which requires the pruning schedule to be periodic and allows for weights pruned erroneously in one cycle to recover in subsequent ones. Experimental results on both linear models and large-scale deep neural networks show that cyclical pruning outperforms existing pruning algorithms, especially at high sparsity ratios. Our approach is easy to tune and can be readily incorporated into existing pruning pipelines to boost performance.
\end{abstract}

\section{Introduction}

The dominant paradigm for training and inference of deep neural networks uses dense parameter 
tensors and hardware optimized for dense computations. However, sparse tensor multiplications 
can be more compute, memory and power efficient, all of which are important considerations for 
low-power mobile devices. To utilize sparsity, we require methods to either train sparse neural networks from scratch or convert existing dense models to sparse ones. Fortunately, it has been shown that pre-trained dense deep neural networks can be easily sparsified using simple heuristics involving magnitude pruning and re-training \cite{han2015learning, zhu2017prune}. One such commonly-used heuristic is gradual pruning, which iteratively prunes weights and follows it with re-training, each time increasing the number of weights pruned. 

On the other hand, recent review papers \cite{gale2019state, blalock2020state} have shown that 
is difficult to improve upon these simple heuristics, and as a result, current state-of-the-art approaches~\cite{gale2019state} still rely on such techniques. In this work, we connect these heuristics to projected gradient descent (PGD), a well-known algorithm for constrained optimization. Similar to gradual pruning, PGD involves alternating between magnitude pruning and 
re-training. However, as we show in the paper, this analogy breaks down for the simple case of pruning a single 
weight. To bridge this gap, we propose \textit{cyclical pruning}, 
a simple strategy that uses a cyclical schedule for pruning rather than a monotonically increasing 
one. Our experimental results show that cyclical pruning outperforms 
gradual pruning across datasets on various models, especially at large sparsity ratios.
This approach does not introduce 
any hard-to-tune hyper-parameters and can be readily incorporated into existing pruning pipelines.

Overall, our contributions are:

\begin{itemize}
    \item We propose \textit{cyclical pruning}, a simple pruning strategy that allows for recovery of previously pruned weights.    
    \item We show that recovery of pruned weights is crucial in the context
    of pruning linear models, especially when the solution space is non-degenerate.
    \item We show improvements over gradual pruning on CIFAR-10 and Imagenet datasets across several models, especially at large sparsity ratios.
\end{itemize}

\section{Related Work}
Pruning in neural networks involves either structured pruning which removes entire neurons, or 
unstructured pruning which removes individual weights. Structured pruning~\cite{li2016pruning, he2017channel, jaderberg2014speeding}, typically does not require any specialized hardware support, as opposed to unstructured pruning which requires explicit support for sparse computations~\cite{choquette2020nvidia, ignatov2018ai}, and is the main focus on this paper.

Methods for unstructured pruning of neural network weights typically rely on magnitude pruning and re-training~\cite{han2015learning, guo2016dynamic, zhu2017prune}, and our paper extends these methods to allow for recovery of pruned weights.
In these works, each layer can either be pruned to the same level of sparsity by applying magnitude pruning 
on each layer separately, or it can be applied once globally. While in this work we use 
local uniform layerwise sparsity, \cite{Kusupati20, azarian2020learned} propose to improve global sparsity by automatically tuning thresholds for magnitude pruning. 

Another orthogonal line of work involves replacing magnitude pruning with alternatives that explicitly consider the impact of pruning on the final loss. To this end, second-order~\cite{lecun1990optimal, hassibi1993second}, and Fisher approximations~\cite{theis2018faster, singh2020woodfisher} of the loss function have been employed. However, recent work~\cite{laurent2021revisiting} has shown that these methods do not necessarily improve upon magnitude pruning, especially when combined with fine-tuning.

Distinct from the approaches considered above, probabilistic approaches to pruning involve approximating 
the original pruning problem via stochastic relaxations~\cite{neklyudov, christosl0, bayesiancompression, vibnet}. These typically involve stochastic optimization over binary gate variables, in addition to the usual optimization 
over weights. However, recent work~\cite{gale2019state} has shown that such techniques often perform on par with, 
simpler magnitude pruning based approaches, which is the focus of this paper.

Also related is the lottery-ticket hypothesis~\cite{frankle2018lottery, lottery_imagenet}, which 
states that there exists a pruning mask for every initialization of a deep model that allow for training only the 
the resulting sparse model from scratch, without the need to alter the pruning mask. While we propose to refine
the mask during pruning using our method, we do not check whether these masks also correspond to lottery tickets,
as this is outside the scope of this work.

Weight recovery has been an important consideration recently in the context of training sparse neural networks from scratch. While \cite{evci2020rigging, jayakumar2020top} use gradient updates to perform weight 
recovery, \cite{dettmers2019sparse} use momentum to do the same. However, both methods place the constraint
that intermediate models obtained during the course of optimization are also sparse, which places heavy 
restrictions on the weight recovery methods. However, no such restrictions apply to our case.
\cite{guo2016dynamic, Lin2020Dynamic} use gradient updates computed on a proxy sparse model by using the straight-through estimator (STE) similar to \cite{wortsman2019discovering} and claim that this can lead to weight recovery. However these methods also use gradual pruning, for which we show that weight recovery is unlikely in practice.

\newcommand{\Y}{\mathbf{y}}
\newcommand{\W}{\mathbf{w}}
\newcommand{\X}{\mathbf{x}}
\newcommand{\Xb}{\mathbf{X}}
\newcommand{\R}{\mathbb{R}}

\section{Methods}\label{sec:methods}
In this section, we discuss existing approaches for unstructured pruning which involve magnitude pruning and re-training,
and introduce the cyclical pruning algorithm. Given the similarity among these approaches, it is helpful to discuss these as instances of a more general framework for pruning, which we call \textit{time-varying projected gradient descent} (TV-PGD), shown in Algorithm \ref{algo}. The distinguishing features of this algorithm when compared to classical PGD are the
usage of an iteration-dependent (or time-varying) sparsity and learning rate function and updating the pruning mask every $\Delta t$ iterations. Here, $\texttt{magprune}(\theta, s(t_i))$ refers to magnitude pruning of $\theta$ with a sparsity ratio of $s(t_i)$, which refers to \textit{global} pruning
in the literature. When this is applied separately layerwise, it is called \textit{local} pruning. Global pruning can result in different 
sparsity rates for every layer, whereas local pruning ensures every layer is pruned to the same sparsity ratio. In this work, we only consider local 
pruning, but these methods equally apply to global pruning.

\newcommand{\N}{\mathbb{N}}

\begin{algorithm}
  \caption{Time-Varying Projected Gradient Descent}\label{algo}
  $\theta \in \R^d$ : model weights, $\ell(\theta) \in \R_+$ : loss function \\
  $T \in \N$ : \# iterations, $\Delta t \in \N$ : pruning interval \\
  $M \in \{0,1\}^d$ : pruning mask, $t_i \in \N_0$: iteration number\\
  $s(t_i) \in [0,1]$ : sparsity function, $\eta(t_i) \in \R_+$ : learning rate  
  
  \begin{algorithmic}[1]
    \Procedure{TV-PGD}{$\theta$}
        \For{$t_i \in [0, T-1]$ iterations}
            \State $\theta \leftarrow \theta - \eta(t_i) ~\nabla_{\theta}\ell(\theta)$\Comment{(S)GD update}
            
            \If{$t_i \bmod \Delta t = 0$} 
            \State $M \leftarrow \texttt{magprune}(\theta, s(t_i))$ \Comment{Get Mask}
            \EndIf
            
            \State $\theta \leftarrow \theta \odot M$\Comment{Prune weights in-place}
        \EndFor
    \EndProcedure
  \end{algorithmic}
\end{algorithm}

\textbf{One-Shot Pruning:} This simple procedure involves two steps: first magnitude pruning the dense model 
according to the target sparsity ratio, and then fine-tuning the resulting sparse model~\cite{han2015learning}. 
This corresponds to TV-PGD 
with $\Delta t > T$ and a sparsity function such that $s(0) = s_t$ equal to the final target sparsity. The learning rate $\eta(t_i)$ is monotonically decreasing in accordance with common training practices in deep learning, except at $t_i=0$, where we have $\eta(0) = 0$.   

\textbf{Gradual Pruning:} This involves pruning with a gradually increasing sparsity schedule, with pruning 
interspersed with fine-tuning. There are two broad variants of this procedure. The first, also called `iterative pruning' \cite{han2015learning} typically performs pruning in few steps (5-10), interspersed with fine-tuning for a large (usually 10+) number of epochs. This corresponds to TV-PGD with linearly increasing $s(t_i) = \frac{s_t ~ t_i}{T}$, $\Delta t \sim 10+$ epochs, and a cyclical learning rate schedule $\eta(t_i)$ such that $\eta(t_i \bmod \Delta t)$ is a monotonically decreasing function according to standard training practices, except for $\eta(t_i \bmod \Delta t = 0) = 0$. Note that one-shot pruning emerges as a special case if $\Delta t > T$.

On the other hand, the `cubic pruning' performs pruning several times (typically 100+), interspersed with a short fine-tuning stage of about few hundred iterations \cite{zhu2017prune}. This corresponds to TV-PGD with a smaller $\Delta t \sim 100$ iterations, and monotonically decreasing $\eta(t_i)$ (as opposed to cyclic) learning rate schedule, and a monotonically increasing cubic sparsity schedule as follows. Here $s_{init}$ is the initial sparsity value and $s_t$ is the target sparsity value.

\begin{equation}
    s(t_i) = s_t + (s_{init} - s_t) \left( 1 - \frac{t_i}{T} \right)^3
    \label{eqn:zg}
\end{equation}

Note that specific implementations of these pruning algorithms may differ slightly from the TV-PGD interpretation, specifically in the usage of in-place pruning which involves directly zero-ing out weights in the weight tensor, but in practice we found no difference in performance between these different variants. 

One characteristic of both one-shot and gradual pruning is their lack of a mechanism for \textit{weight recovery}, i.e., the ability of pruned weights to be recovered in future steps, which we define below.

\begin{defn}
(Weight Recovery) is said to have occurred in TV-PGD for some weight $j$ if at any two iterations $t_1, t_2$ such that $t_2 > t_1$, we have the pruning masks $M_j(t_1) = 0$ and $M_j(t_2) = 1$.
\end{defn}

Intuitively, we expect weight recovery to help in cases where identification of the correct weights to prune are critical, and where one-shot magnitude pruning does not identify these. In such cases, weight recovery can 
help correct mistakes made by magnitude pruning, and allow pruning of different weights in subsequent steps. However if correct identification of weights to prune does not matter, then we do not expect weight recovery to help. We further elaborate upon this in \S \ref{sec:linear_pruning}. We observe that in TV-PGD, weight recovery can only occur if a magnitude pruning step immediately follows a gradient update step, resulting in a dense weight tensor. 

For one-shot pruning and iterative pruning, we observe that weight recovery is impossible as the magnitude pruning step never occurs immediately after performing a dense SGD update step. For cubic pruning, while weight recovery is technically possible, we found that it is highly improbable in practice. We hypothesize that this happens because weight recovery requires the magnitude of weights after a single update for some pruned weight at $j$ to be larger than that of the smallest unpruned weight, which is improbable owing to usage of relatively small and monotonically decreasing learning 
rates $\eta(t_i)$ used for fine-tuning. In other words, we require $(\eta(t_i) \nabla_{\theta} \ell(\theta))_j^2 > 
\min_{k, \theta_k > 0} \theta_k^2$, for pruned weights $\theta$, which is difficult to satisfy when $\eta(t_i)$ is small, and training diverges if $\eta(t_i)$ is set high. We thus require a procedure that can 
reliably grow back weights that have been pruned previously, and for this purpose, we introduce a simple strategy called \textit{cyclical pruning}.

\begin{figure}
    \centering
    \includegraphics[width=10cm]{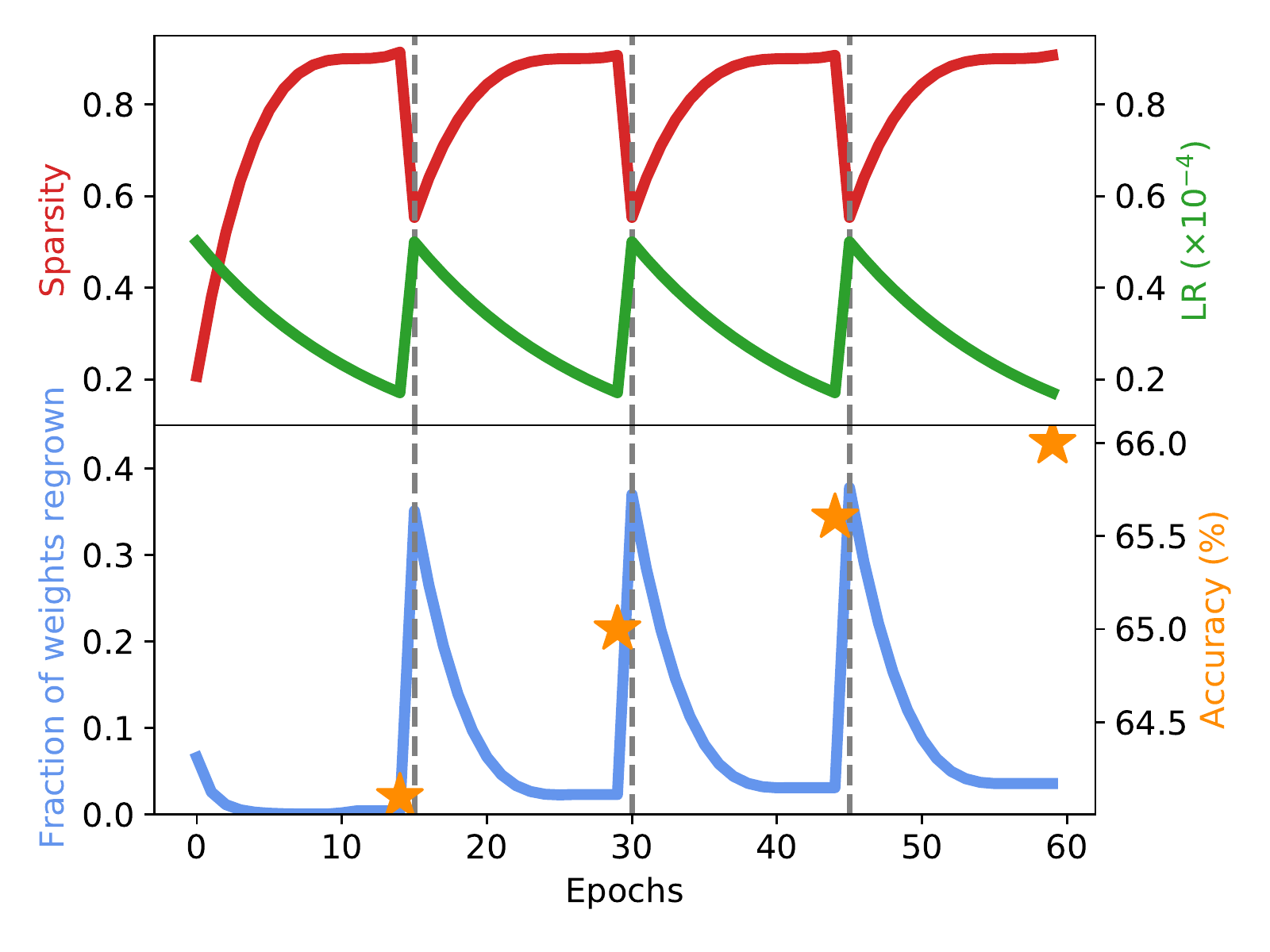}
    \caption{Illustration of cyclical pruning for ResNet18 on Imagenet with $k=4$ cycles, which specifies both the \textcolor{Red}{sparsity} and the \textcolor{OliveGreen}{learning rate} schedules. This procedure results in an increase in the \textcolor{CornflowerBlue}{number of weights recovered} during pruning, and also a corresponding increase in \textcolor{orange}{pruned model accuracy} across cycles. Note that a weight is considered regrown if it was pruned during any previous training step, but unpruned at the current step.}
    \label{fig:main}
\end{figure}

\textbf{Cyclical Pruning:} We propose to perform pruning with a cyclical pruning schedule rather than a 
monotonically increasing one. Specifically, we divide the overall pruning schedule into $k$ cycles, and within
each cycle the pruning schedule is monotonically increasing. This corresponds to TV-PGD with a periodic 
$s(t_i)$, where each cycle has a monotonically increasing $s(t_i \bmod T/k)$, and a periodic $\eta(t_i)$ 
corresponding to a monotonically decreasing $\eta(t_i \bmod T/k)$, and $\Delta t \sim 100$ iterations. We observe that weight recovery always occurs here due to the periodic resetting of the sparsity rates, i.e., $s(t_2) < s(t_1)$ for $t_2 > t_1$ ensures weight recovery. Further, the recovered weights are likely to recover on par with unpruned weights due to the periodic re-setting of learning rates,
as this allows for sufficient updates for important weights to recover. In practice, we use the same per-cycle sparsity 
schedule as in equation \ref{eqn:zg}, and use a different value of
$s_{init}$ for the first cycle ($s_{init} = 0$), and subsequent cycles ($s_{init} \sim 0.5 s_t$), although
we did not find this to be crucial.

\textbf{PGD Pruning: } A classical method to perform constrained optimization is projected gradient
descent (PGD), which in this case corresponds to TV-PGD with $\Delta t = 1$ and constant functions $\eta(t_i), 
s(t_i)$. We notice that similar to cyclical pruning, PGD also allows for weight recovery at all steps. However,
in practice we find this to be ineffective for pruning deep neural networks, owing to the possible instability 
caused by pruning at every iteration, and the improbability of recovery due to $( \eta(t_i) \nabla_{\theta} \ell(\theta) )^2_j$ being small. Hence practical considerations such as mini-batching and usage of relatively
small, monotonically decreasing learning rates reduce its effectiveness in practice. 

Thus weight recovery is a distinguishing feature of both cyclical pruning and PGD. In the next section, we take a 
closer look at weight recovery in PGD while pruning linear models, where practical considerations of training deep models do not apply.

\section{Is Weight Recovery Necessary?}\label{sec:linear_pruning}
In this section we study the importance of weight recovery in the simple case of sparse linear
regression with a single pruned weight. Formally, let $\hat{\Y} = \W^\top \Xb$, where $\Y \in \R^n, \Xb \in \R^{d \times n}, \W \in \R^{d}$. Also assume that outputs are generated from an underlying sparse vector, i.e, $\Y = \alpha^\top \Xb$, 
where $\| \alpha \|_0 = d-1$, and $\alpha_{c} =0$ for some index $c \in \{1,...,d\}$. Also assume that the problem is over-parameterized ($d > n$). Here, we wish to solve the following.

\begin{align}
    \W^* = \arg \min_{\W, \| \W \|_0 \leq d-1} \| \Y - \W^\top \Xb \|^2
    \label{prob:splinreg}
\end{align}

In general, sparse linear regression is NP-hard \cite{natarajan1995sparse}, however if $\Xb$ satisfies the Restricted Isometry Property (RIP) \cite{candes2006robust}, then efficient polynomial time solutions are known to exist (i.e., PGD) \cite{blumensath2008iterative}. Note that specifics of the RIP condition are not relevant to our discussion here, and we refer interested readers to \cite{jain2017nonconvex}. In practice, it is easy to construct approximately RIP matrices by sampling matrix entries from a scaled unit normal distribution \cite{baraniuk2008simple}. As a result, we henceforth assume that $\Xb$ satisfies RIP, and begin by making the following observation.

\begin{obs}
\label{obs:equivalence}
For problem \ref{prob:splinreg}, gradual pruning (with $\eta(0) = 0$) is equivalent to one-shot pruning, and
cyclical pruning (with $\Delta t = 1$) is equivalent to projected gradient descent.
\end{obs}

This is true because for pruning a single weight, $s(t_i)$ for any \textit{strictly} monotonically increasing schedule reduces to a step function, and the cyclical schedule $s(t_i)$ reduces a constant function $s(t_i) = s_{t}$ for $\Delta t = 1$. Further, the usage of mini-batches is unnecessary here, and we use full-batch gradient descent instead. We thus only 
study one-shot pruning and PGD as proxies for studying gradual and cyclical pruning respectively. For PGD, strong recovery guarantees hold for the sparse linear regression problem under some regularity conditions~\cite{blumensath2008iterative, jain2017nonconvex}. Note that discussion of these conditions is out of
scope for this paper.


Unfortunately, such guarantees do not hold for one-shot pruning.
This is easy to see by applying one-shot pruning on randomly initialized weights. 
Here the first step involves magnitude pruning which effectively prunes a random weight, and the probability of pruning the correct index $c$ at initialization is only $\frac{1}{d}$. The second step involves 
re-training, which cannot change the pruned weight. Thus with overwhelming probability $(\frac{d-1}{d})$, random initialization followed by one-shot pruning fails to recover $\alpha$. This simple counter-example illustrates
why such recovery guarantees cannot hold for one-shot pruning.

However this analysis does not reflect standard practice in pruning where one-shot pruning is typically after 
dense training, and not on randomly initialized weights. For the linear case, dense training corresponds to solving 
an unconstrained version of equation \ref{prob:splinreg}, which is solved via regularized least-squares method. 

Assuming some $\lambda > 0$ for regularized least-squares, let $\mathbb{A} = (\Xb^\top \Xb + \lambda I)^{-1} \Xb^\top \Xb$, then it is easy to see that the least-squares solution is $\W = \mathbb{A} \alpha$. We can use this to construct a problem (i.e, pick $\alpha$) such that for some index $c$ with $\alpha_c = 0$, we have $c \neq \arg\min_i \W_i^2$. This ensures that magnitude pruning performed on the least squares solution $\W$ does not select the correct index $c$. One such choice of $\alpha$ is as follows: $\alpha_i = 
\begin{cases}
\mathbb{A}[c,i], & i \neq c \\
0, & i = c
\end{cases}$.

This ensures that $\W = \mathbb{A} \alpha$ has a large magnitude on the $c^{th}$ co-ordinate, causing magnitude pruning 
to select an incorrect index, which we call the \textbf{adversarial} choice 
of $\alpha$. 

Running simulations on this problem with $d=5, n=4, c=3$, and sampling $10^4$ different RIP matrices $\Xb$, we find that magnitude pruning succeeds in picking the correct index $c$ only $\sim 3\%$ of the time.
We also empirically observe that if we set $d >> n$, then this probability tends to zero. We summarize the results
of the simulations in the following statement.

\begin{obs}
We find empirically that it is possible to choose solutions $\alpha$ for problem \ref{prob:splinreg} such that dense training followed by one-shot pruning fails to recover $\alpha$ with high probability.
\end{obs}

This shows that even in the realistic setting of one-shot pruning applied after dense training, there exists
problems such that one-shot pruning fails to select the correct index. Note that PGD is immune to this in principle 
as the recovery guarantees are independent of initialization.

\subsection{When does PGD fail?}

Having considered instances where one-shot pruning fails, we now ask the converse question: when does PGD fail? 
The regularity conditions of PGD recovery \cite{jain2017nonconvex} indicate that this can happen for severely 
over-parameterized problems, i.e., $n << d$. In this case, the linear system $\Y = \W^\top\Xb$ not only maintains 
infinitely many dense solutions for $\W$, but also has multiple sparse solutions. To see this, we rewrite the linear 
system as $(\W - \alpha)^\top \Xb = 0$, which implies that $\W - \alpha$ lies in the left-nullspace of $\Xb$. As an 
example, assume that the dimensionality of the left null-space is $d_{null} = 2$ for problem dimension 
$d = 5, n = 3$ and the task of pruning a single weight. Let $v_1, v_2 \in \R^d$ be the basis vectors of the 
left null-space. Then, it is possible to find $\gamma_1, \gamma_2 \in \R$ such that the system of equations $\gamma_1 \times v_{1,i} + \gamma_2 \times v_{2,i} = \W_i - \alpha_i $ and $ \gamma_1 \times v_{1,j} + \gamma_2 \times v_{2,j} = \W_j - \alpha_j$ is satisfied for any two arbitrary indices $i,j \in \{1,..,d\}$, assuming that the determinant of $\begin{bmatrix} 
v_{1,i} & v_{2,i}\\
v_{1,j} & v_{2,j}
\end{bmatrix}$ is nonzero. 
In particular, we can set $i = c$, such that $\alpha_c = \W_c = \W_j = 0$ and the
system still maintains a solution. This implies that we have a valid solution $\| \W \|_0 = 3$ which is sparser than the ground truth solution $\| \alpha \|_0 = 4$. A simple generalization states that a $(n,d)$ over-parameterized ($d > n$) sparse linear regression problem has at least a $d-n$-dimensional left null-space, which contains 
at most ${d \choose d-n}$ number of $n-$sparse solution vectors.

\begin{table}
    \centering
    \vspace{-0.5cm}
    \caption{Simulation results for 100 runs of sparse linear regression performed after regularized least squares on a
    problem with $d=5$. We observe that when the problem is severely over-parameterized, both PGD and one-shot pruning perform similarly, while PGD outperforms one-shot pruning in other scenarios. This property also holds for larger image datasets and deep neural networks (see \S \ref{subsec:sota}).}
    \label{tab:simulation}
    \begin{tabular}{cccc}
        \parbox{1.5cm}{\centering \# samples ($n$)} & \parbox{1.5cm}{\centering Choice of $\alpha$} & \parbox{1.5cm}{\centering Prob. of one-shot recovery} & \parbox{1.5cm}{\centering Prob. of PGD recovery}\\ \toprule
        \multirow{2}{*}{\parbox{2cm}{ \centering 2 ($n << d$)}} & random & \textbf{0.33} & \textbf{0.32} \\
         & adversarial & \textbf{0.12} & \textbf{0.11} \\
        \multirow{2}{*}{\parbox{2cm}{ \centering 4 ($n < d$)}} & random & 0.22 & \textbf{0.36} \\
         & adversarial & 0.02 & \textbf{0.23} \\
        \multirow{2}{*}{\parbox{2cm}{ \centering 10 ($n > d$)}} & random & 0.35 & \textbf{0.64} \\
         & adversarial & 0.04 & \textbf{0.44} \\
         \bottomrule
    \end{tabular}
\end{table}

Thus when the problem is severely over-parameterized, we can always find $n$-sparse solutions $\W$ irrespective 
of the sparsity of the ground truth solution $\alpha$. If $\| \alpha \|_0 > n$, then sparse recovery is not 
possible and we are able to prune more weights ($n$) than the solution $\alpha$ requires. This impossibility of 
sparse recovery renders PGD ineffective, making it no more effective than simple one-shot pruning.


To verify this, we run simulations on problems with different number of samples $n$ and different choices for $\alpha$, chosen either randomly or adversarially. The results of this simulation are given 
in Table \ref{tab:simulation}. Note that in all cases we perform either one-shot pruning or PGD on the regularized least squares solution. We observe that one-shot pruning recovers the optimal solution at the 
same rate as PGD for the severely over-parameterized case, but PGD outperforms one-shot pruning in other cases. 
We attribute the less than perfect solution recovery of PGD to the non-adherance to the regularity conditions 
for both PGD and gaurantees for gaussian matrices to be RIP \footnote{Specifically, we do not ensure the RIP with an 
order $ >3(d-1)$ is maintained, as this would not allow for $(d-1)$-sparsity that applies to Observation \ref{obs:equivalence}}.

Summarizing this section, we first find that one-shot pruning can fail for linear models, 
while PGD is more likely to converge to the correct solution, both in theory and 
practice. We next find that when the problem is severely over-parameterized, we can prune more weights than the 
solution requires and thus sparse recovery is not possible, which renders PGD no more effective than one-shot pruning. We stress here that the analysis done here is limited to the case of linear models, as notions such as RIP do not apply to non-linear regression using deep neural networks.
However as we shall see in the next section, this behaviour of 
one-shot pruning and PGD on linear models carries over to their respective proxies, i.e., gradual and cyclical pruning applied to deep neural networks trained on large image datasets. This is because the underlying principle is same in both cases, i.e., when the decision of which weight to prune is not important, then weight recovery does not help.

\section{Experiments}\label{sec:expts}

In this section, we show detailed experimental results and ablation studies examining various aspects of cyclical pruning. Our experiments are done using the Pytorch framework~\cite{pytorch_paper}, and are organized as follows.
In \S~\ref{subsec:sota}, we compare cyclic pruning with state-of-the-art pruning algorithms. Here, we show results on CIFAR10 \cite{cifar_dataset} and Imagenet \cite{imagenet_cvpr09} datasets across various models. 
In \S~\ref{subsec:schedule}, we perform controlled ablation experiments to study the impact
of the sparsity and learning rate schedules, and the behaviour of the algorithm across different cycles.

\newcommand{\acc}[2]{#1 \scriptsize{$\pm$ #2}}
\newcommand{\accb}[2]{\textbf{#1} \scriptsize{$\pm$ #2}}

\begin{table*}[!h]
	\centering
	\caption{%
    Accuracy (\%) after pruning various models on the CIFAR-10 dataset, with gradual pruning and one-shot 
    pruning run for 100 epochs, and cyclical pruning for 5 cycles of 20 epochs each. We observe that cyclical pruning offers an advantage primarily at larger sparsity values, while being competitive at smaller values, in accordance with the theory in \S \ref{sec:linear_pruning}.
    }
    \label{tab:cifar10}
	\begin{tabular}{lccccc}
		
		&& \multicolumn{3}{c}{Methods} 	& 	\\ \cmidrule{3-5}
		\parbox{1cm}{Model}  & \parbox{1cm}{\centering Baseline}	 & \parbox{2.5cm}{\centering One-Shot Pruning \cite{han2015learning}}	& \parbox{2.5cm}{\centering Gradual Pruning \cite{zhu2017prune}} 	& \parbox{2.5cm}{\centering Cyclical Pruning \\ (Ours)}	& 	\parbox{1cm}{\centering Pruning ratio}	 \\ \toprule
        
        \multirow{4}{*}{ResNet-56}          & \multirow{4}{*}{$93.28$}          	& \accb{92.35}{0.1} 		& \accb{92.44}{0.0} 		&                                       \accb{92.41}{0.1}		& 90\%      \\ 
				            			    & 										& \acc{90.85}{0.0}		& \accb{91.69}{0.1} 		& \accb{91.90}{0.2}		& 95\%      \\ 
				            			    & 										& \acc{79.22}{0.0}		& \acc{89.57}{0.1} 		& \accb{90.54}{0.0}		& 98\%      \\ 
				            			    & 										& \acc{58.03}{0.3}		& \acc{68.20}{0.1} 		& \accb{70.99}{0.3}		& 99\%      \\ \midrule
        \multirow{4}{*}{Mobilenet}          & \multirow{4}{*}{$89.75$}          	& \accb{90.22}{0.2} 	& \accb{90.25}{0.0} 	&                                       \acc{89.83}{0.1}		& 70\%      \\ 
				            			    & 										& \acc{88.44}{0.0}		& \accb{89.49}{0.2} 		& \accb{89.37}{0.0}		& 80\%      \\ 
				            			    & 										& \acc{84.99}{0.3}		& \acc{85.51}{0.7} 		& \accb{86.99}{0.3}		& 90\%      \\ 
				            			    & 										& \acc{75.05}{1.0}		& \acc{73.42}{0.0} 		& \accb{79.07}{0.6}		& 95\%      \\
	\bottomrule	
	\end{tabular}%
\end{table*}

\begin{table*}[!h]
	\centering
	\caption{%
    Accuracy (\%) after pruning various models on the Imagenet dataset, with gradual pruning and one-shot 
    pruning run for 60 epochs, and cyclical pruning for 3 cycles of 20 epochs each. We observe that cyclical pruning offers an advantage primarily at larger sparsity values, while being competitive at smaller values, in accordance with the theory in \S \ref{sec:linear_pruning}.
    }
    \label{tab:imagenet}
	\begin{tabular}{lccccc}
		
		&& \multicolumn{3}{c}{Methods} 	& 	\\ \cmidrule{3-5}
		\parbox{1cm}{Model}  & \parbox{1cm}{\centering Baseline}	 & \parbox{2.5cm}{\centering One-Shot Pruning\\ \cite{han2015learning}}	& \parbox{2.5cm}{\centering Gradual Pruning\\ \cite{zhu2017prune} } 	& \parbox{2.5cm}{\centering Cyclical Pruning \\ (Ours)}	& 	\parbox{1cm}{\centering Pruning ratio}	 \\ \toprule
        
        \multirow{4}{*}{ResNet18}           & \multirow{4}{*}{69.7}
				            			    & 										 $\mathbf{69.9}$		& $\mathbf{69.9}$ 		& $69.6$		& 60\%  \\ 
				            			    & 										& $69.2$		& $69.2$ 		& $\mathbf{69.4}$		& 70\%  \\ 
				            			    & 										& $68.2$		& $67.8$ 		& $\mathbf{68.3}$		& 80\%  \\ 
				            			    & 										& $63.5$		& $63.6$ 		& $\mathbf{64.9}$		& 90\%  \\\midrule
		\multirow{5}{*}{ResNet50}           & \multirow{5}{*}{76.16}  	& $75.9$ 		& $\mathbf{76.1}$ 		& $75.8$		& 60\%  \\ 
				            			    & 										& $\mathbf{75.9}$		& $75.8$ 		& $75.7$		& 70\%  \\ 
				            			    & 										& $\mathbf{75.4}$		& $74.9$ 		& $\mathbf{75.3}$		& 80\%  \\ 
				            			    & 										& $72.8$		& $71.9$ 		& $\mathbf{73.3}$		& 90\%  \\
				            			    & 										& $67.1$		& $64.7$ 		& $\mathbf{68.7}$		& 95\%  \\ \midrule		            			    
		\multirow{5}{*}{EfficientNet}        & \multirow{5}{*}{$74.8$} 	& $73.9$		& $74.0$		& $\mathbf{74.1}$	    & 40\%  \\
				            				& 										& $73.2$		& $73.2$ 		& $\mathbf{73.4}$		& 50\%  \\ 
				            			    & 										& $71.2$		& $71.8$ 		& $\mathbf{72.4}$		& 60\%  \\ 
				            			    & 										& $68.0$		& $68.2$ 		& $\mathbf{69.9}$		& 70\%  \\
				            			    & 										& $65.1$		& $65.2$ 		& $\mathbf{67.5}$		& 75\%  \\
				            			    \midrule
		\multirow{4}{*}{MobilenetV2}       & \multirow{4}{*}{$71.7$} 	& $70.8$		& $\mathbf{70.9}$		& $69.8$	    & 40\%  \\
				            				& 										& $67.6$		& $69.8$ 		& $\mathbf{70.1}$		& 50\%  \\ 
				            			    & 										& $66.7$		& $67.6$ 		& $\mathbf{68.4}$		& 60\%  \\ 
				            			    & 										& $61.3$		& $62.7$ 		& $\mathbf{64.4}$		& 70\%  \\
				            			     									
	\bottomrule	
	\end{tabular}%
\end{table*}

\begin{table*}[!h]
\caption{Informal comparison of cyclical pruning with published results on Resnet50 trained on Imagenet. We only compare with methods that use local sparsity. For cyclical pruning, we start from a dense pre-trained ResNet-50 trained for 90 epochs, and use a cycle length of 20 epochs for pruning. Thus the total number of epochs corresponds to 130 epochs = 90 + 2 cycles $\times$ 20 epochs, and similarly for 110 \& 150 epochs. Note that 110 epochs of cyclical pruning corresponds to one cycle, thus being identical to gradual pruning. Other methods in literature train from scratch.}
    \label{tab:imagenet_sota}
\centering
 \begin{tabular}{l c c c c} 
 Method & Pruning ratio & Dense Baseline & Pruned & Difference\\ 
 \toprule
 SNFS~(100 epochs) \cite{dettmers2019sparse} & 72.4\% & 75.95 & 74.59 & -1.36 \\
 DPF ~(90 epochs) \cite{Lin2020Dynamic} & 73.5\% & 75.95 & 75.48 & -0.47 \\
  Cyclical Pruning (110 epochs, Ours) & 73.5\% & 76.16 & 75.46 & -0.7\\
 {Cyclical Pruning (130 epochs, Ours)} & 73.5\% & 76.16 & 75.84 & \textbf{-0.32}\\ 
 Cyclical Pruning (150 epochs, Ours) & 73.5\% & 76.16 & 76.00 & \textbf{-0.15}\\ \midrule
 
 SNFS~(100 epochs) \cite{dettmers2019sparse} & 82.0\% & 75.95 & 72.65 & -3.30 \\
 RigL (100 epochs) \cite{evci2020rigging} & 80.0\% & 76.80 & 74.60 & -2.2\\
 RigL (500 epochs) \cite{evci2020rigging} & 80.0\% & 76.80 & 76.60 & \textbf{-0.2} \\
 DPF~(90 epochs) \cite{Lin2020Dynamic} & 82.6\% & 75.95 & 74.55 & -1.44 \\
 Gradual Pruning (100 epochs) \cite{gale2019state} & 80.0\% & 76.69 & 75.58 & -1.11\\
 Cyclical Pruning (110 epochs, Ours) & 82.6\% & 76.16 & 74.65 & -1.51\\ 
 {Cyclical Pruning (130 epochs, Ours)} & 82.6\% & 76.16 & 75.29 & \textbf{-0.87}\\ 
 {Cyclical Pruning (150 epochs, Ours)} & 82.6\% & 76.16 & 75.40 & \textbf{-0.75}\\
 \bottomrule
 \end{tabular}
\end{table*}

\subsection{Comparison with Gradual Pruning}\label{subsec:sota}

Here we shall compare cyclical pruning with two baselines: one-shot 
pruning and gradual pruning. These are in accordance with the best practices 
suggested by \cite{blalock2020state}. In particular, we consider overall 8 
architecture-dataset pairs with modern architectures, we report values along 
the trade-off curve for all methods, we compare different methods using an identical
model, library and optimizer setup, and we report standard deviations whenever possible.
We do not report explicit compression ratio and speedup as we use local layerwise sparsity,
and hence these numbers are identical across different methods for the same sparsity ratio.

We first discuss results on the CIFAR10 dataset 
across two models\footnote{Results on additional CIFAR10 / CIFAR100 models are provided in the supplementary.}. 
We present results at sparsity ratios from $90\%$ to $99\%$, for 
Resnet56, and $70\%$ to $95\%$ for Mobilenet owing to the compact nature
of this model. For rigorous comparisons, we perform pruning from the same baseline 
model in all cases, and allow each method the same amount of computation. Specifically, 
we train for 100 epochs for one-shot pruning and gradual pruning, and use 20 epochs with 5 cycles for cyclical pruning. 
We use SGD with momentum as our optimizer, and use the same learning rate schedules in all cases within a single cycle, 
i.e., we start fine-tuning with a learning rate of $1e-2$ and drop it to $1e-3$ after completing $75\%$ of the 
allocated epochs, and use a batch size of $256$. For the cyclical sparsity, the allocated epochs corresponds to the 
number of epochs for a single cycle, in this case being $20$. Our experimental results
in Table \ref{tab:cifar10} shows that cyclical pruning outperforms gradual pruning, especially at high sparsity 
ratios. This aligns perfectly with the observations made for pruning of linear models, where 
PGD showed no benefits for the case of severe over-parameterization, which in this case corresponds to pruning
with smaller sparsity ratios.

We also show experimental results on Imagenet in Table \ref{tab:imagenet}, where we show results on four pre-trained models with varying sparsity levels. In this case, for one-shot pruning and gradual pruning we allow 
60 epochs of training, and use 20 epochs with 3 cycles for cyclical pruning. We use an exponential learning 
rate schedule. The learning rate is always decreased at the halfway mark, i.e., by a factor of 10 every 10 epochs for the cyclical pruning and every 30 epochs for gradual pruning. We use starting learning rate $1e-4$ and Adam optimizer~\cite{kingma2014adam}, with a batch size of $64$ for all the experiments in Table \ref{tab:imagenet}. The experiments show that cyclical pruning outperforms gradual pruning for higher compression ratios for all the models. This further supports the observations made earlier for pruning of linear models.

In addition to this, we also make informal comparisons with other reported results in literature in Table 
\ref{tab:imagenet_sota}. Note that comparisons with reported results are not recommended practice~\cite{blalock2020state}, and this only provides an approximate indication of the relative performance of 
different methods. Furthermore, other methods in literature are trained sparse networks from scratch, whereas we prune pre-trained models. Although it is possible to apply our method to train from scratch as well, we do not do this here due to lack of resources to tune hyper-parameters for full Imagenet training. Also here we only make comparisons with methods which use local sparsity, similar to us. For instance, 
\cite{evci2020rigging, gale2019state} also provide pruning results with global sparsity, but we do not compare against those.\footnote{For results of \cite{gale2019state} see: \url{https://bit.ly/39KSC6Z}}. 
The highest claimed performance in literature that we are aware of is by \cite{evci2020rigging} who run 500 
epochs of training on Imagenet, which is several times larger than our computational budget. For cyclical pruning 
of Resnet 50 in Table \ref{tab:imagenet_sota} we use SGD with momentum with a
learning rate of $1e-2$ and momentum $0.9$, instead of Adam which was used to obtain results of Table 
\ref{tab:imagenet}.

\subsection{Ablation Experiments}\label{subsec:schedule}

\begin{table*}[!h]
    \centering
    \caption{Comparison of different per-cycle pruning schedules used with cyclical pruning, on Resnet20 / CIFAR10 @ 99\% sparsity. `Mask distance' refers to a Jaccard distance computed between a given mask and the mask obtained after
    the first cycle. We observe that while cubic schedule performs the best, we observe that in all cases the accuracy increases and the pruning mask changes every cycle. This provides evidence for weight recovery in cyclical pruning. We also observe that simply training longer using cyclical learning rates is insufficient.}
    \label{tab:schedule_ablation}
    \setlength\tabcolsep{2pt}
    \begin{tabular}{cc ccccc}
        & Cycle \# & 1 & 2 & 3 & 4 & 5 \\ \toprule
        \multirow{2}{*}{Cubic schedule} & Accuracy (\%) & \acc{62.74}{0.2} &  \acc{65.89}{0.6} &  \acc{66.89}{0.7} & \acc{67.23}{0.6} & \acc{67.56}{0.7}  \\[0.2cm] 
                                                 & \parbox{2.5cm}{\centering Mask distance} & 0 & 0.38 & 0.49 & 0.55 & 0.58 \\ \midrule
        \multirow{2}{*}{Linear schedule} & Accuracy (\%) & \acc{47.63}{1.2} & \acc{52.035}{0.9} & \acc{56.33}{0.9} & \acc{57.65}{1.2} & \acc{58.27}{1.2} \\[0.2cm] 
                                                 & \parbox{2.5cm}{\centering Mask distance} & 0 & 0.48 & 0.58 & 0.62 & 0.64 \\ \midrule
        \multirow{2}{*}{Step schedule} & Accuracy (\%) & \acc{39.88}{0.5} & \acc{47.47}{0.4} & \acc{50.94}{0.4} & \acc{53.64}{0.2} & \acc{55.36}{0.6} \\[0.2cm] 
                                                 & \parbox{2.5cm}{\centering Mask distance} & 0 & 0.46 & 0.59 & 0.65 & 0.68  \\ \midrule 
        \multirow{2}{*}{\parbox{2.5cm}{\centering Finetune with\\ cyclical learning rates}} & Accuracy (\%) & \acc{63.01}{0.0} & \acc{63.73}{0.1} & \acc{64.11}{0.2} & \acc{63.8}{0.2} & \acc{64.06}{0.2} \\[0.2cm] 
                                                 & \parbox{2.5cm}{\centering Mask distance} & 0 & 0 & 0 & 0 & 0 \\[0.1cm]\bottomrule 
    \end{tabular}
\end{table*}

Here we perform controlled experiments 
to understand the behaviour of cyclical pruning across
successive cycles. First, we consider the effect of the pruning schedule within a 
single cycle by comparing the evolution of the model upon using a cubic schedule, with 
that of linear and step schedules. Here, step schedule corresponds to one-shot pruning, 
which can be thought of as using heaviside step function for the sparsity schedule.
In all cases, each cycle consists of 20 epochs, and within each cycle, 16 epochs are 
used for alternating pruning and re-training, and the final 4 epochs consist of purely 
fine-tuning of the sparse model. For step pruning, we prune at the halfway mark, i.e., 
at 10 epochs. The results shown in Table \ref{tab:schedule_ablation} indicate that regardless 
of the sparsity schedule, cyclical pruning always shows accuracy improvements over successive cycles,
with cubic pruning performing the best overall. Further, we also compute the Jaccard distance of 
pruning masks obtained at the end of cycles to understand the extend of mask evolution. Table 
\ref{tab:schedule_ablation} also shows that the mask changes drastically across cycles in all cases,
thus confirming our hypothesis that cyclical pruning allows for correction of erroneously pruned weights.

Second, we decouple the effect of cyclical pruning with cyclical learning rates and show that the improvement
across cycles is precisely due to the pruning schedule and not due the learning rate schedule. To test this, 
we run a control experiment referred to as `Finetune with cyclical learning rates' in Table \ref{tab:schedule_ablation},
where the first cycle is identical to cyclical pruning, and in the subsequent cycles only 
fine-tuning is performed for the obtained sparse model without additional pruning. For this fine-tuning, we maintain the learning rate schedule used for cyclical sparsity. We observe that 
increasing the number of fine-tuning epochs does lead to improved accuracy across cycles as expected, but not as much as that 
obtained for cyclical sparsity. As expected, fine-tuning also cannot allow any changes in the pruning mask which leads to zero Jaccard 
distances.

Overall, the experiments in Table \ref{tab:schedule_ablation} here show that, (1) Accuracy improves across rounds in cyclical pruning regardless of the 
sparsity schedule, and weight recovery indeed takes place as indicated by the Jaccard distances. (2) Cubic pruning outperforms linear and one-shot (step) pruning. (3) Improved performance of cyclical pruning is \textbf{not} explained by longer training, 
as shown by the comparison with cyclical learning rates.

\section{Discussion}
In this work, we introduce \textit{cyclical pruning}, a simple strategy that allows for 
recovery of erroneously pruned weights, leading to an improved sparsity-accuracy trade-off across 
various datasets and models. The cyclical paradigm can be used in conjunction with any per-cycle 
sparsity schedule. In addition, the cycle-wise accuracy improvements also show that cyclical sparsity 
can also be used to improve performance of existing sparse models obtained via any other method. Our 
theory and experiments reveal that cyclical pruning offers an advantage primarily for pruning with large 
sparsity ratios, when the solution space is not degenerate and the choice of weights to prune is critical.
This method introduces only two additional hyper-parameters, the number of cycles $k$ and the initial sparsity $s_{init}$
for subsequent cycles. We found that it is generally beneficial to keep the number of cycles, and the number of
epochs per cycle to be as large as possible within the computational budget, and we show in the supplementary that the 
choice of $s_{init}$ is not crucial. These indicate that the method is also easy to tune.

Future work involves understanding why cubic pruning works well, and whether it is 
possible to use cyclical pruning in a more efficient manner that decouples its dependence on cubic pruning. 
Our unified view of pruning methods via TV-PGD also leads to a natural open problem: how do we optimally set $(s(t_i), \eta(t_i), \Delta t)$ in TV-PGD to guarantee convergence for neural network pruning? 

\bibliography{references}

\begin{thebibliography}{10}

\bibitem{han2015learning}
Song Han, Jeff Pool, John Tran, and William Dally.
\newblock Learning both weights and connections for efficient neural network.
\newblock {\em Advances in neural information processing systems},
  28:1135--1143, 2015.

\bibitem{zhu2017prune}
Michael~H. Zhu and Suyog Gupta.
\newblock To prune, or not to prune: Exploring the efficacy of pruning for
  model compression.
\newblock 2018.

\bibitem{gale2019state}
Trevor Gale, Erich Elsen, and Sara Hooker.
\newblock The state of sparsity in deep neural networks.
\newblock {\em arXiv preprint arXiv:1902.09574}, 2019.

\bibitem{blalock2020state}
Davis Blalock, Jose Javier~Gonzalez Ortiz, Jonathan Frankle, and John Guttag.
\newblock What is the state of neural network pruning?
\newblock {\em arXiv preprint arXiv:2003.03033}, 2020.

\bibitem{li2016pruning}
Hao Li, Asim Kadav, Igor Durdanovic, Hanan Samet, and Hans~Peter Graf.
\newblock Pruning filters for efficient convnets.
\newblock {\em International Conference on Learning Representations (ICLR)},
  2017.

\bibitem{he2017channel}
Yihui He, Xiangyu Zhang, and Jian Sun.
\newblock Channel pruning for accelerating very deep neural networks.
\newblock In {\em Proceedings of the IEEE International Conference on Computer
  Vision}, pages 1389--1397, 2017.

\bibitem{jaderberg2014speeding}
Max Jaderberg, Andrea Vedaldi, and Andrew Zisserman.
\newblock Speeding up convolutional neural networks with low rank expansions.
\newblock {\em arXiv preprint arXiv:1405.3866}, 2014.

\bibitem{choquette2020nvidia}
Jack Choquette and Wish Gandhi.
\newblock Nvidia a100 gpu: Performance \& innovation for gpu computing.
\newblock In {\em 2020 IEEE Hot Chips 32 Symposium (HCS)}, pages 1--43. IEEE
  Computer Society, 2020.

\bibitem{ignatov2018ai}
Andrey Ignatov, Radu Timofte, William Chou, Ke~Wang, Max Wu, Tim Hartley, and
  Luc Van~Gool.
\newblock Ai benchmark: Running deep neural networks on android smartphones.
\newblock In {\em Proceedings of the European Conference on Computer Vision
  (ECCV) Workshops}, pages 0--0, 2018.

\bibitem{guo2016dynamic}
Yiwen Guo, Anbang Yao, and Yurong Chen.
\newblock Dynamic network surgery for efficient dnns.
\newblock {\em Advances in Neural Information Processing Systems}, 2016.

\bibitem{Kusupati20}
Aditya Kusupati, Vivek Ramanujan, Raghav Somani, Mitchell Wortsman, Prateek
  Jain, Sham Kakade, and Ali Farhadi.
\newblock Soft threshold weight reparameterization for learnable sparsity.
\newblock In {\em Proceedings of the International Conference on Machine
  Learning}, July 2020.

\bibitem{azarian2020learned}
Kambiz Azarian, Yash Bhalgat, Jinwon Lee, and Tijmen Blankevoort.
\newblock Learned threshold pruning.
\newblock {\em arXiv preprint arXiv:2003.00075}, 2020.

\bibitem{lecun1990optimal}
Yann LeCun, John~S Denker, and Sara~A Solla.
\newblock Optimal brain damage.
\newblock In {\em Advances in neural information processing systems}, pages
  598--605, 1990.

\bibitem{hassibi1993second}
Babak Hassibi and David~G Stork.
\newblock Second order derivatives for network pruning: Optimal brain surgeon.
\newblock In {\em Advances in neural information processing systems}, pages
  164--171, 1993.

\bibitem{theis2018faster}
Lucas Theis, Iryna Korshunova, Alykhan Tejani, and Ferenc Husz{\'a}r.
\newblock Faster gaze prediction with dense networks and fisher pruning.
\newblock {\em arXiv preprint arXiv:1801.05787}, 2018.

\bibitem{singh2020woodfisher}
Sidak~Pal Singh and Dan Alistarh.
\newblock Woodfisher: Efficient second-order approximations for model
  compression.
\newblock {\em Advances in Neural Information Processing Systems}, 2020.

\bibitem{laurent2021revisiting}
C{\'e}sar Laurent, Camille Ballas, Thomas George, Pascal Vincent, and Nicolas
  Ballas.
\newblock Revisiting loss modelling for unstructured pruning, 2021.

\bibitem{neklyudov}
Kirill Neklyudov, Dmitry Molchanov, Arsenii Ashukha, and Dmitry~P Vetrov.
\newblock Structured bayesian pruning via log-normal multiplicative noise.
\newblock In I.~Guyon, U.~V. Luxburg, S.~Bengio, H.~Wallach, R.~Fergus,
  S.~Vishwanathan, and R.~Garnett, editors, {\em Advances in Neural Information
  Processing Systems 30}, pages 6775--6784. Curran Associates, Inc., 2017.

\bibitem{christosl0}
Christos Louizos, Max Welling, and Diederik~P. Kingma.
\newblock Learning sparse neural networks through l0 regularization.
\newblock In {\em International Conference on Learning Representations}, 2018.

\bibitem{bayesiancompression}
Christos Louizos, Karen Ullrich, and Max Welling.
\newblock Bayesian compression for deep learning.
\newblock In I.~Guyon, U.~V. Luxburg, S.~Bengio, H.~Wallach, R.~Fergus,
  S.~Vishwanathan, and R.~Garnett, editors, {\em Advances in Neural Information
  Processing Systems 30}, pages 3288--3298. Curran Associates, Inc., 2017.

\bibitem{vibnet}
Bin Dai, Chen Zhu, and David Wipf.
\newblock Compressing neural networks using the variational information
  bottleneck.
\newblock {\em arXiv preprint arXiv:1802.10399}, 2018.

\bibitem{frankle2018lottery}
Jonathan Frankle and Michael Carbin.
\newblock The lottery ticket hypothesis: Finding sparse, trainable neural
  networks.
\newblock {\em arXiv preprint arXiv:1803.03635}, 2018.

\bibitem{lottery_imagenet}
Jonathan Frankle, Gintare~Karolina Dziugaite, Daniel~M Roy, and Michael Carbin.
\newblock The lottery ticket hypothesis at scale.
\newblock {\em arXiv preprint arXiv:1903.01611}, 8, 2019.

\bibitem{evci2020rigging}
Utku Evci, Trevor Gale, Jacob Menick, Pablo~Samuel Castro, and Erich Elsen.
\newblock Rigging the lottery: Making all tickets winners.
\newblock In {\em International Conference on Machine Learning}, pages
  2943--2952. PMLR, 2020.

\bibitem{jayakumar2020top}
Siddhant Jayakumar, Razvan Pascanu, Jack Rae, Simon Osindero, and Erich Elsen.
\newblock Top-kast: Top-k always sparse training.
\newblock In H.~Larochelle, M.~Ranzato, R.~Hadsell, M.~F. Balcan, and H.~Lin,
  editors, {\em Advances in Neural Information Processing Systems}, volume~33,
  pages 20744--20754. Curran Associates, Inc., 2020.

\bibitem{dettmers2019sparse}
Tim Dettmers and Luke Zettlemoyer.
\newblock Sparse networks from scratch: Faster training without losing
  performance.
\newblock {\em arXiv preprint arXiv:1907.04840}, 2019.

\bibitem{Lin2020Dynamic}
Tao Lin, Sebastian~U. Stich, Luis Barba, Daniil Dmitriev, and Martin Jaggi.
\newblock Dynamic model pruning with feedback.
\newblock In {\em International Conference on Learning Representations}, 2020.

\bibitem{wortsman2019discovering}
Mitchell Wortsman, Ali Farhadi, and Mohammad Rastegari.
\newblock Discovering neural wirings.
\newblock In {\em Advances in Neural Information Processing Systems}, 2019.

\bibitem{natarajan1995sparse}
Balas~Kausik Natarajan.
\newblock Sparse approximate solutions to linear systems.
\newblock {\em SIAM journal on computing}, 24(2):227--234, 1995.

\bibitem{candes2006robust}
Emmanuel~J Cand{\`e}s, Justin Romberg, and Terence Tao.
\newblock Robust uncertainty principles: Exact signal reconstruction from
  highly incomplete frequency information.
\newblock {\em IEEE Transactions on information theory}, 52(2):489--509, 2006.

\bibitem{blumensath2008iterative}
T~Blumensath and ME~Davies.
\newblock Iterative hard thresholding for compressed sensing.
\newblock 2008.

\bibitem{jain2017nonconvex}
Prateek Jain and Purushottam Kar.
\newblock Non-convex optimization for machine learning.
\newblock {\em Foundations and Trends® in Machine Learning}, 10(3-4):142--363,
  2017.

\bibitem{baraniuk2008simple}
Richard Baraniuk, Mark Davenport, Ronald DeVore, and Michael Wakin.
\newblock A simple proof of the restricted isometry property for random
  matrices.
\newblock {\em Constructive Approximation}, 28(3):253--263, 2008.

\bibitem{pytorch_paper}
Adam Paszke, Sam Gross, Francisco Massa, Adam Lerer, James Bradbury, Gregory
  Chanan, Trevor Killeen, Zeming Lin, Natalia Gimelshein, Luca Antiga, Alban
  Desmaison, Andreas Kopf, Edward Yang, Zachary DeVito, Martin Raison, Alykhan
  Tejani, Sasank Chilamkurthy, Benoit Steiner, Lu~Fang, Junjie Bai, and Soumith
  Chintala.
\newblock {PyTorch: An Imperative Style, High-Performance Deep Learning
  Library}.
\newblock In H~Wallach, H~Larochelle, A~Beygelzimer,
  F~d$\backslash$textquotesingle Alch{\'{e}}-Buc, E~Fox, and R~Garnett,
  editors, {\em Advances in Neural Information Processing Systems}, volume~32,
  pages 8026--8037. Curran Associates, Inc., 2019.

\bibitem{cifar_dataset}
Alex Krizhevsky and Geoffrey Hinton.
\newblock Learning multiple layers of features from tiny images.
\newblock 2009.

\bibitem{imagenet_cvpr09}
J.~Deng, W.~Dong, R.~Socher, L.-J. Li, K.~Li, and L.~Fei-Fei.
\newblock {ImageNet: A Large-Scale Hierarchical Image Database}.
\newblock In {\em CVPR09}, 2009.

\bibitem{kingma2014adam}
Diederik~P Kingma and Jimmy Ba.
\newblock Adam: A method for stochastic optimization.
\newblock {\em arXiv preprint arXiv:1412.6980}, 2014.

\end{thebibliography}
\bibliographystyle{unsrt}
\clearpage

\section*{Appendix}

\begin{figure*}[ph]
\centering
\includegraphics[width=10cm]{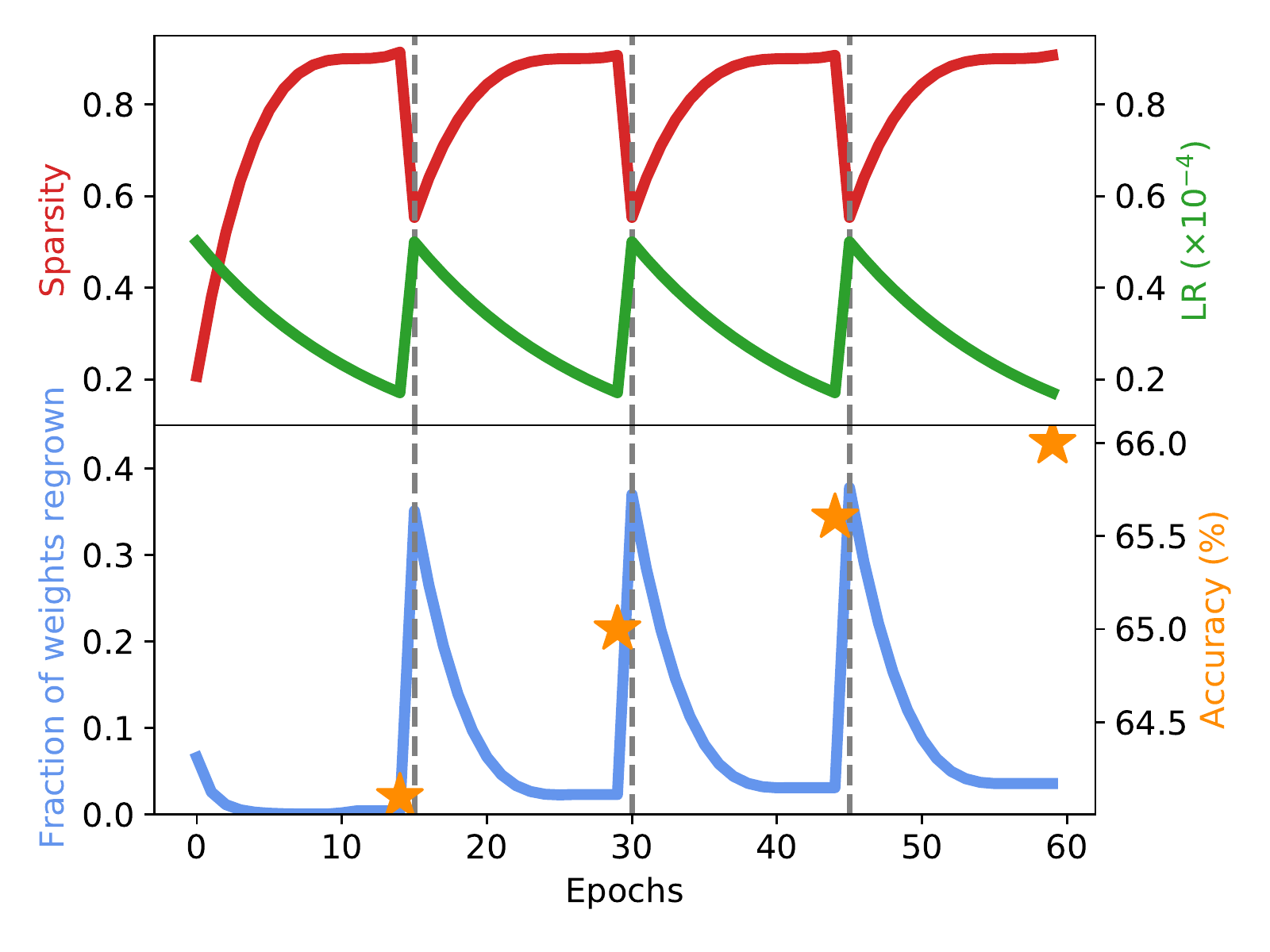}

\caption{Evidence for weight recovery on Resnet18 trained on ImageNet. The model is pruned to 90\% sparsity for 4 cycles of 15 epochs each. In plot (a) we count the number of regrown weights. A weight is considered to be regrown if it is not pruned at the current iteration, but it was pruned at least once in any previous iteration. The plots from the top to bottom show the sparsity, the fraction of weights regrown, and the learning rate respectively. In plot (b) we show the fraction of weights regrown by the end of each cycle versus validation accuracy. The plot indicates that the improvement of the accuracy over the cycles corresponds to an increasing amount of regrown weights.}
\label{fig:ablation_regrown_weights}
\end{figure*}

\subsection{Evidence of Weight Recovery in Cyclical Pruning (\& its Absence in Gradual Pruning)}

In Figure \ref{fig:ablation_regrown_weights} we analyze the effect of the cyclical pruning schedule on weight 
recovery during pruning, on a Resnet18 model trained on Imagenet. This model is pruned to 4 cycles of 15 epochs 
each, with the rest of the hyper-parameters the same as that described in the main text. We visualize two aspects 
of this process. First we plot the fraction of weights recovered across 
epochs. Here we observe that the first cycle of cyclical sparsity is identical to gradual pruning \cite{zhu2017prune}, where we observe little to no weights recovered at the end of the first cycle. At
the beginning of the second cycle, we note that a large number of weight recover owing to the resetting in 
sparsity values and learning rates. This leads to a substantial number of weights at the end of the cycle 
as well. Next, we observe that the fraction of weights regrown increase
with every cycle, also leading to a corresponding increase in accuracy. Note that here we use $s_{init} = 0.5$ 
from the second cycle onwards, where $s_{init}$ is used in gradual pruning below.

\begin{equation}
    s(t_i) = s_t + (s_{init} - s_t) \left( 1 - \frac{t_i}{T} \right)^3
    \label{eqn:zg}
\end{equation}

\subsection{Comparison with Gradual Pruning on CIFAR10 / CIFAR100}

Here we compare the performance of cyclical pruning with gradual and one-shot pruning, similar to that 
shown in the main text. The results in \ref{tab:cifar100} show trends similar to that seen for CIFAR10 
and Imagenet, namely that cyclical pruning is effective for large sparsity ratios and is competitive 
with gradual pruning in small ones. The experimental settings remain identical to those used for CIFAR10.

\begin{table*}[!h]
	\centering
	\caption{%
    Accuracy (\%) after pruning various models on the CIFAR-10 dataset, with gradual pruning and one-shot 
    pruning run for 100 epochs, and cyclical pruning for 5 cycles of 20 epochs each. We observe that cyclical pruning offers an advantage primarily at larger sparsity values, while being competitive at smaller values, in accordance with the theory in the main text.
    }
    \label{tab:cifar10}
	\begin{tabular}{lccccc}
		
		&& \multicolumn{3}{c}{Methods} 	& 	\\ \cmidrule{3-5}
		\parbox{1cm}{Model}  & \parbox{1cm}{\centering Baseline}	 & \parbox{2.5cm}{\centering One-Shot Pruning \cite{han2015learning}}	& \parbox{2.5cm}{\centering Gradual Pruning \cite{zhu2017prune}} 	& \parbox{2.5cm}{\centering Cyclical Pruning \\ (Ours)}	& 	\parbox{1cm}{\centering Pruning ratio}	 \\ \toprule
        
        \multirow{4}{*}{ResNet-20}          & \multirow{4}{*}{$92.24$}          	& \acc{90.10}{0.2} 		& \accb{90.78}{0.4} 		&                                       \accb{90.90}{0.1}		& 90\%      \\ 
				            			    & 										& \acc{86.52}{0.9}		& \accb{89.14}{0.2} 	& \accb{89.29}{0.2}		& 95\%      \\ 
				            			    & 										& \acc{78.55}{0.2}		& \acc{83.44}{0.1} 		& \accb{85.79}{0.1}		& 98\%      \\ 
				            			    & 										& \acc{33.43}{0.0}		& \acc{50.77}{2.7} 		& \accb{66.04}{2.4}		& 99\%      \\ \midrule
        \multirow{4}{*}{ResNet-56}          & \multirow{4}{*}{$93.28$}          	& \accb{92.35}{0.1} 		& \accb{92.44}{0.0} 		&                                       \accb{92.41}{0.1}		& 90\%      \\ 
				            			    & 										& \acc{90.85}{0.0}		& \accb{91.69}{0.1} 		& \accb{91.90}{0.2}		& 95\%      \\ 
				            			    & 										& \acc{79.22}{0.0}		& \acc{89.57}{0.1} 		& \accb{90.54}{0.0}		& 98\%      \\ 
				            			    & 										& \acc{58.03}{0.3}		& \acc{68.20}{0.1} 		& \accb{70.99}{0.3}		& 99\%      \\ \midrule
        \multirow{4}{*}{VGG-14}             & \multirow{4}{*}{$93.57$}          	& \acc{92.98}{0.1} 		& \accb{93.25}{0.1} 		&                                       \acc{93.03}{0.1}		& 90\%      \\ 
				            			    & 										& \accb{92.17}{0.1}		& \accb{92.36}{0.2} 	& \accb{92.47}{0.0}		& 95\%      \\ 
				            			    & 										& \acc{89.29}{0.1}		& \acc{90.64}{0.3} 		& \accb{91.71}{0.1}		& 98\%      \\ 
				            			    & 										& \acc{85.75}{0.0}		& \acc{88.59}{0.0} 		& \accb{89.59}{0.7}		& 99\%      \\ \midrule
        \multirow{4}{*}{Mobilenet}          & \multirow{4}{*}{$89.75$}          	& \accb{90.22}{0.2} 	& \accb{90.25}{0.0} 	&                                       \acc{89.83}{0.1}		& 70\%      \\ 
				            			    & 										& \acc{88.44}{0.0}		& \accb{89.49}{0.2} 		& \accb{89.37}{0.0}		& 80\%      \\ 
				            			    & 										& \acc{84.99}{0.3}		& \acc{85.51}{0.7} 		& \accb{86.99}{0.3}		& 90\%      \\ 
				            			    & 										& \acc{75.05}{1.0}		& \acc{73.42}{0.0} 		& \accb{79.07}{0.6}		& 95\%      \\
	\bottomrule	
	\end{tabular}%
\end{table*}

\begin{table*}[h]
	\centering
	\caption{%
    Accuracy (\%) of various models on the CIFAR-100 dataset, with gradual pruning and one-shot 
    pruning run for 100 epochs, and cyclical pruning for 5 cycles of 20 epochs each. We observe that cyclical pruning offers an advantage primarily at larger sparsity values, while being competitive at smaller values, in accordance with our theory, and the CIFAR10 and Imagenet experiments in the main text.
    }
    \label{tab:cifar100}
	\begin{tabular}{lccccc}
		
		&& \multicolumn{3}{c}{Methods} 	& 	\\ \cmidrule{3-5}
		\parbox{1cm}{Model}  & \parbox{1cm}{\centering Baseline}	 & \parbox{2.5cm}{\centering One-Shot Pruning \\\cite{han2015learning}}	& \parbox{2.5cm}{\centering Gradual Pruning\\ \cite{zhu2017prune}} 	& \parbox{2.5cm}{\centering Cyclical Pruning \\ (Ours)}	& 	\parbox{1cm}{\centering Pruning ratio}	 \\ \midrule
        
        \multirow{4}{*}{ResNet-20}          & \multirow{4}{*}{$72.53$}      & \acc{70.57}{0.0} 		& \accb{71.20}{0.0} 		&  \accb{71.44}{0.2}		& 90\%      \\ 
				            			    & 								& \acc{67.28}{0.1}		& \acc{69.32}{0.0} 		& \accb{70.53}{0.1}		& 95\%      \\ 
				            			    & 								& \acc{57.29}{0.1}		& \acc{62.56}{0.0} 		& \accb{66.81}{0.0}		& 98\%      \\ 
				            			    & 								& \acc{43.23}{0.0}		& \acc{48.68}{2.1} 		& \accb{60.63}{0.2}		& 99\%      \\ \midrule
        \multirow{4}{*}{ResNet-56}          & \multirow{4}{*}{$76.78$}      & \acc{74.48}{0.1} 		& \accb{75.19}{0.2} 		& \accb{74.97}{0.0}		& 90\%      \\ 
				            			    & 								& \acc{72.03}{0.1}		& \accb{73.41}{0.4} 		& \accb{73.86}{0.1}		& 95\%      \\ 
				            			    & 								& \acc{68.48}{0.0}		& \acc{70.62}{0.3} 		& \accb{72.46}{0.0}		& 98\%      \\ 
				            			    & 								& \acc{46.73}{2.7}		& \accb{56.62}{1.9} 		& \accb{56.79}{0.0}		& 99\%      \\ \midrule
        \multirow{4}{*}{VGG-14}             & \multirow{4}{*}{$74.35$}      & \accb{72.36}{0.1} 	& \accb{72.49}{0.2} 		& \accb{72.22}{0.1}		& 90\%      \\ 
				            			    & 								& \acc{70.09}{0.2}		& \acc{71.06}{0.0} 		& \accb{71.54}{0.4}		& 95\%      \\ 
				            			    & 								& \acc{64.52}{0.1}		& \accb{67.60}{0.0} 		& \accb{67.61}{0.0}		& 98\%      \\ 
				            			    & 								& \acc{49.27}{0.0}		& \acc{59.47}{0.0} 		& \accb{64.76}{0.0}		& 99\%      \\ \midrule
        \multirow{4}{*}{Mobilenet}          & \multirow{4}{*}{$63.67$}      & \accb{64.87}{0.0} 	& \accb{64.91}{0.5} 		&  \accb{64.43}{0.1}		& 70\%      \\ 
				            			    & 								& \accb{64.04}{0.1}		& \accb{64.37}{0.2} 		& \acc{63.91}{0.2}		& 80\%      \\ 
				            			    & 								& \acc{60.72}{0.0}		& \acc{58.48}{0.8} 		& \accb{62.08}{0.4}		& 90\%      \\ 
				            			    & 								& \acc{49.59}{0.2}		& \acc{40.25}{0.3} 		& \accb{55.03}{1.0}		& 95\%      \\ 
	\bottomrule	
	\end{tabular}%
\end{table*}

\subsection{Impact of $s_{init}$}

Here we shall analyse the impact of $s_{init}$ parameter used in Equation \ref{eqn:zg}, used for cyclical pruning
from the second cycle onwards. We run an ablation study on the Resnet20 model on CIFAR10 across 4 values of $s_{init}$ in Table \ref{tab:start_sparsity}, and find that across the values tried, the performance of the 
pruned models are not affected. This indicates that the setting of $s_{init}$ is not crucial to the pruning
performance.

\begin{table*}[h]
    \centering
    \caption{Accuracy(\%) of model trained with different $s_{init}$ values in cubic scheduling (ref. Eq \ref{eqn:zg}) used from cycle 2 onwards, for Resnet20 / CIFAR10 @ 98\% sparsity. We observe that this value is not crucial to tune.}
    \label{tab:start_sparsity}
    \begin{tabular}{c ccccc}
        \parbox{2.5cm}{\centering Cycle \# $\rightarrow$ \\ Start Sparsity $\downarrow$ } & 1 & 2 & 3 & 4 & 5\\ \midrule
        {0}  & \acc{81.64}{0.2} &  \acc{83.85}{0.1} &  \acc{84.64}{0.4} & \acc{85.51}{0.1} & \acc{85.76}{0.0}  \\[0.2cm] 
        {25\%} & \acc{81.41}{0.2} &  \acc{83.99}{0.1} &  \acc{84.58}{0.0} & \acc{85.76}{0.1} & \acc{85.90}{0.4}  \\[0.2cm] 
        {50\%} & \acc{81.04}{0.2} &  \acc{83.92}{0.1} &  \acc{84.81}{0.1} & \acc{85.24}{0.1} & \acc{85.58}{0.1}  \\[0.2cm] 
        {75\%} & \acc{81.42}{0.1} &  \acc{84.01}{0.3} &  \acc{84.77}{0.3} & \acc{85.39}{0.1} & \acc{85.69}{0.2}  \\\bottomrule 
    \end{tabular}
\end{table*}

\subsection{Comparison with "Gradual Pruning with recovery"}
Here we consider a variant of gradual pruning which allows for weight recovery in between successive pruning 
steps. Specifically, after every pruning step, we apply gradient updates and allow the resulting model to be dense until the next pruning step, typically occurring after 100 iterations. This corresponds to a modification of TV-PGD, where the \texttt{prune weights in-place} step in moved inside the \texttt{if} block. 
In Table \ref{tab:tijmen_zg} we see this variant of gradient pruning performs on par with usual gradual pruning, 
and does not outperform cyclical pruning. This implies that allowing more recovery in this fashion is still 
insufficient.

\begin{table}
	\centering
	\caption{%
    Comparison of accuracies (\%) of cyclical pruning with a variant of gradual pruning which allows for weight recovery. We observe that cyclical pruning still 
    performs better as gradual pruning is still unable to perform weight recovery for the final pruning steps. (Move to supplementary)
    }
    \label{tab:tijmen_zg}
	\begin{tabular}{cccc}
		
		\parbox{1.5cm}{\centering Gradual Pruning} 	
		& \parbox{1.5cm}{\centering Cyclical Pruning \\ (Ours) }	
		& \parbox{1.5cm}{\centering Gradual Pruning with Recovery} 
		&	\parbox{1.5cm}{\centering Pruning ratio}	 \\ \midrule
        
         \accb{90.78}{0.4} 	& \accb{90.90}{0.1}	 & \acc{90.66}{0.0} & 90\%      \\ 
		\accb{89.14}{0.2} 	& \accb{89.29}{0.2}	 & \acc{88.74}{0.2}	& 95\%      \\ 
		\acc{83.44}{0.1} 		& \accb{85.79}{0.1}	 & \acc{83.36}{0.8}	& 98\%      \\ 
		 \acc{50.77}{2.7} 		& \accb{66.04}{2.4}	 & \acc{56.94}{2.5}	& 99\%      \\ 
				            			    \bottomrule	
	\end{tabular}%
\end{table}

\subsection{Training Cyclical Pruning from Scratch}

In this section, we shall evaluate the performance of cyclical pruning when combined with model training from scratch. Here we are faced with the problem that the learning rate schedules prescribed for cyclical pruning are not suitable for training from scratch, which requires a specific monotonically decreasing schedule. As an example, typical Imagenet training runs for 90 epochs, with the learning rate decaying by 10 times at epochs 30 and 60. These do not allow for increasing learning rates. 

To get around this problem, here we run a baseline with pruning after 60 epochs of Imagenet training (after which there are no learning rate decreases), and hence is a partially pre-trained model. We here run cyclical pruning with a cycle width of 15 epochs instead of 20 epochs used in the paper, and present the results in Table \ref{tab:scratch}.

These numbers are on an average 1-1.2\% lower than their cyclical pruning counterparts in Table 4 in the main paper, which prune pre-trained models. Thus our experiments do not yet show the effectiveness of cyclical pruning on training from scratch, particularly due to their incompatibility with typical learning rate schedules. However, we believe an extensive hyper-parameter sweep over possible cyclical learning rate schedules on Imagenet may be more effective, which is outside our computational budget.

\begin{table}
    \centering
    \caption{Results of training models with cyclical pruning from scratch.}
    \label{tab:scratch}
    \begin{tabular}{c c c}
         \toprule
         \textbf{Sparsity} & \textbf{Pre-training + Pruning Epochs} & \textbf{Accuracy (\%)}\\
         \midrule
         73.5 &	60 + 15 = 75 &	74.61 \\
73.5 &	60 + 15*2 = 90 &	74.88 \\
73.5 &	60 + 15*3 = 105 &	74.81 \\
82.6 &	60 + 15 = 75 &	73.75 \\
82.6 &	60 + 15*2 = 90 &	74.25 \\
82.6 &	60 + 15*3 = 105 &	74.38 \\
    \bottomrule
    \end{tabular}
\end{table}

\end{document}